\pdfoutput=1

\documentclass[11pt]{article}

\usepackage{acl}

\usepackage{times}
\usepackage{latexsym}

\usepackage[T1]{fontenc}

\usepackage[utf8]{inputenc}

\usepackage{microtype}

\title{Methods for Estimating and Improving Robustness of Language Models}

\author{Michal Štefánik \\ 
  Faculty of Informatics, Masaryk University \\
  \texttt{stefanik.m@mail.muni.cz}}
\begin{document}
\maketitle
\begin{abstract}
Despite their outstanding performance, large language models (LLMs) suffer notorious flaws related to their preference for simple, surface-level textual relations over full semantic complexity of the problem. This proposal investigates a common denominator of this problem in their weak ability to generalise outside of the training domain. We survey diverse research directions providing estimations of model generalisation ability and find that incorporating some of these measures in the training objectives leads to enhanced distributional robustness of neural models. Based on these findings, we present future research directions towards enhancing the robustness of LLMs.
\end{abstract}

\section{Introduction}

The advances in language processing that we observe in recent years, mostly led by the instances of large language models (LLMs) based on the transformer architecture \cite{attention} raise a deserved attention of the scientific community. We find studies concluding that LLMs fine-tuned for a specific task can align with, or even outperform human accuracy on complex tasks such as question answering \cite{rajpurkar-etal-2016-squad}, paraphrase identification \cite{bowman-etal-2015-snli}, machine translation \cite{cho} and others.

In contrast, critical studies demonstrate that many of the models reaching a state-of-the-art on a given task perform poorly on data sets drawn from different distribution(s).
This is due to various reasons, such as training data set biases including spurious linguistic correlations \cite{McCoy2019RightFT}, different text stylistics or typos \cite{nise_breaks_translation}, where a broad preference of LLMs towards fitting non-representative, yet easy-to-learn surface-level relations cause them to under-perform even shallow networks \cite{bojanowski2016enriching}. A~lack of generalisation can also be caused by procedural reasons, such as training process instability, causing a convergence to local minima of distinct generalisation quality \cite{berts_dont_generalize}. 
Low robustness of the consequential model towards out-of-distribution (OOD) samples limits their practical usability to the samples drawn from the training distribution, which is often impossible to ensure. 

Despite that the complex language models strike an impression of a black-box, an extensive branch of research demonstrated that internal representations of LLMs correspond well to a human taxonomy in terms of morphological and syntactic decomposition \cite{what_does_bert_look_at}, or that the depth of the internal representation correlates well with the complexity of the problem as perceived by humans \cite{nlp_pipeline}. 

The reported agility support the central presumption of this proposal; that LLMs can avoid the problems mentioned above under additional \emph{regularisation}. We argue that such regularisation could also strenghten the implicit property of LLMs learning compositional language features and thus enhance an \textit{interpretability} of their decision-making.

In this proposal, we survey literature from the broader area of neural networks for the reasons for better generalisation of the neural model. We find that many measures reported to correlate well with model's OOD performance can also enhance neural model generalisation when utilised within the model's training objective, as regularisers, or additional components of the training cost function. Inspired by this finding, this proposal outlines a path towards identification and utilisation of generalisation measures aimed to enhance robustness of LLMs towards distribution shift.

\begin{description}
\item[RQ1:] ``Can we \emph{estimate} the performance of LLMs on data from OOD, without a collection of annotated data or expert feedback?''
\end{description}

\begin{description}
\item[RQ2:] ``Can we \emph{adjust} the process of training LLMs to perform \emph{better} on OOD samples?''
\end{description}

In Section \ref{sec:evaluation} we survey the studies aiming to estimate robustness of neural models with no restrictions on a domain of application. Subsequently, in Section~\ref{sec:ood-learning}, we survey the training techniques reported to enhance the robustness of the trained model. Based on these findings, in Section~\ref{ch:objectives} we identify promising directions and respective challenges specific for estimating (§\ref{ref:evaluative_measures_objective}) and enhancing (§\ref{sec:objectives_objective}) the robustness of LLMs.

\subsection{Applicability}

This proposal grounds the notion of model generalisation to its ability to perform well on samples drawn from distributions different than the training distribution (OOD). 
In this context, the term of a \textit{distribution}, used interchangeably with \textit{domain}, is commonly described by a specific shared property, such as topic, style, genre, or linguistic register~\cite{ramponi-plank-2020-neural}. 

This proposal focuses on distributional robustness in two branches of applications of current LLMs: \textit{generative tasks}, where the problem is to generate a sequence of tokens, and \textit{discriminative tasks}, where the task is to infer a discrete decision for each token or a sequence of tokens. Generative tasks include summarization, dialogue generation or machine translation, while discriminative tasks include classification, extractive question answering or named entity recognition.

In both cases, we propose to estimate the impact of given adjustment on model generalisation by measuring a difference in the model's performance on a set of distinct OOD domains. 
We note that such estimation is still only a pointwise estimation of model generalisation as some properties of the domains drawn for evaluation remain uncontrolled.

\section{Background}

\subsection{Estimating Model Robustness (RQ1)}
\label{sec:evaluation}

Having a set of true labels for some set of OOD samples $X_t$ of target domain(s) $D_t$, the robustness of the model $M$ can be estimated using standard qualitative measures, such as accuracy. This raises questions about the representativeness of the draw of $X_t$: do these cover \emph{all} the domains of application of $M$, and are these domains accurately weighted in evaluation?

The problem is circumvented by generalisation measures based on \emph{latent properties} of $M$, that do not require any labelled data of $D_t$. However, such an approach might come at the price of accuracy: according to \citet{Jiang2020FantasticGM}, the Spearman's rank correlation of any unsupervised measure with out-of-distribution accuracy does not exceed $0.5$ on average. The accuracy of the estimator improves using supervised approaches \cite{stefanik-etal-2021-regressive}, but these already require some labelled data.

The situation presents a common dilemma in robustness evaluation: Ground-truth evaluation must involve a representative selection of test data. This problem can be avoided with unsupervised estimations based on the model properties, but such proxies are burdened by a certain level of inaccuracy. In the following sections, we review the measures introduced directly for evaluating model generalisation (§\ref{sec:generalisation}) and for estimating model's expected output quality (§\ref{sec:quality_estimation}), more commonly used in NLP.

\subsubsection{Generalisation Measures}
\label{sec:generalisation}

Traditionally, the ability of neural networks to generalise was related to the measures of their \emph{capacity}, where the lower capacity might imply the lower \emph{generalisation gap} \cite{Jiang2020FantasticGM}, i.e. a drop of performance under \emph{distribution shift}. The capacity can be quantified in terms of \emph{complexity} given by a number of model parameters, expressive power or others. A standard example of such a measure is a degree of a polynomial; the higher the degree, the better is the fit, but it comes at the price of generalisation loss. This group of measures is referred to as Vapnik–Chervonenkis dimension (VC-dimension), introduced by \citet{vapnik99}.

A large body of work aims to find such VC-dimensions that correspond well with OOD performance even with modern, over-parametrised networks.
For instance, norm-based approaches \cite{pmlr-v40-Neyshabur15} propose to use the $p$-norms used in regularisation of the training as the anchor value of generalisation and support this in theory by connecting such measure with a limitation of network capacity. \citet{Bartlett2017SpectrallynormalizedMB} conclude that a \emph{spectral complexity} measure, that is inferred from eigenvalues of a matrix of the network weights, can be used as one of such complexity measures. 

A collateral line of work, starting with \citet{705570} show that \emph{generalisation bounds}, denoting a range of expected performance of the given model on an arbitrary test set, can be provably associated with \emph{VC-bounds}.
\citet{pmlr-v65-harvey17a} show that the \emph{tightness} of such bounds for a linear subset of networks can be theoretically found. 
Furthermore, \citet{Dziugaite2017ComputingNG} propose a method to \emph{optimize} \emph{PAC-Bayesian bounds}, optimising the model for as tight bounds as possible.

Despite these proofs, error bounds based on VC-dimensions remain \emph{vacuous} in practice \cite{Dziugaite2017ComputingNG, Jiang2020FantasticGM}: such estimates of OOD performance are too wide to be used in practice. Additionally, it is now widely observed \cite{Novak2018SensitivityAG, Neyshabur2015InSO}, that in practice, an effect of over-parametrisation is in contrast with traditional VC-dimension theory and in multiple cases, over-parametrisation leads to \emph{better} reported generalisation \cite{neyshabur2018the}.

Existing work attempts to ground \emph{error bounds} in the underlying causal model that \emph{describes} the target domains of interest. \citet{Meinshausen2018CAUSALITYFA} introduces a term of \emph{Structural equation model} (SEM) defining the causal interventions consistent with a given \textit{world} and relates domain generalisation to the model's robustness to the \textit{interventions} defined by such SEM. Additionally, given that SEM produces a class of distributions $\mathcal{Q}$, a model $M$ robust on $\mathcal{Q}$ is a \emph{causal inference model} for $\mathcal{Q}$, connecting distributional robustness to a \emph{weak form} of causal inference \cite{Dziugaite2020InSO}. Similarly, \citet{Bhlmann2018InvarianceCA} ascribes the ability of causal inference on $\mathcal{Q}$ to any model whose representation is invariant to any domain $D \in \mathcal{Q}$ and proposes a method of selecting a subset of \emph{invariant features} that picks such subset of attributes from a given set.

Practical observations of errors suggest that empirical \emph{error bounds} are in fact significantly tighter than what can be proven in theory. \citet{Dziugaite2020InSO} locate all bounds between the two extremes: theoretically-supported, yet vacuous bounds of methods based solely on the model property (\emph{VC-bounds}) or behaviour (\emph{PAC-Bayesian bounds}) and empirical, yet strictly data- and model-dependent evaluation on sample set(s) $X_{t} \in D_t$.

\subsubsection{Quality Estimation}
\label{sec:quality_estimation}

\emph{Quality estimation} (QE) measure predicts model output quality in the absence of ground-truth reference \cite{fomicheva-etal-2020-unsupervised}. 
Although not commonly used in this manner, QE measures also reflect on model robustness, making this branch of research applicable for OOD performance estimation (\textbf{RQ1}).

A significant line of work grounds quality estimation in model \emph{confidence}, which can be estimated using Bayesian networks \cite{10.5555/143221} where standard \emph{scalar} weights of the network are replaced with random variables, modelling the output distribution. This approach is accurate but not computationally feasible for larger networks. A~branch of work \emph{approximates} parametric distributions \cite{NIPS2011_7eb3c8be, Tran2019BayesianLA} making such uncertainty estimation practically feasible. 

Model uncertainty can also be computed by ensembling variations of a given model in multiple trials, commonly referred to as \emph{Monte Carlo} (MC) methods. Monte Carlo dropout \cite{MCDropout} applies dropout on inference randomly among multiple inference trials yielding an estimation of the distribution of network output, based on which the uncertainty is approximated. \citet{Lee2015WhyMH} build such ensembles of estimators using \emph{bagging}, i.e. training the ensembled models on different train sub-sets.

Model-variational methods fit well into the central \emph{PAC-Bayesian} theory \cite{10.1145/800057.808710}, stating that if the error of the classifier can be bound, then also a performance of an ensemble of such classifiers can be upper-bound with arbitrarily-small bound $\epsilon$ \cite{Guedj2019APO}.

Confidence estimation can be utilised in enhanced model robustness, where prediction confidence is used as a regularizer of the main objective; in augmentation \cite{Szegedy2014IntriguingPO}, confidence calibration \cite{Gong2021ConfidenceCF}, or in a training for consistency \cite{xie2019unsupervised}.

\citet{Jiang2020FantasticGM} propose to measure a regularisation decay of the weights, together with a measure of \emph{sharpness}, reflecting on a volume of change in the model evaluation when the limited surrounding of the learnt parameter space minima is permuted \cite{Keskar2017OnLT}. Another introduced measure reflects a \emph{variance of gradients} measured on a train set after a first training iteration. This work is the first large-scale study evaluating correlation of selected generalisation measures with true OOD performance and concludes that the mentioned sharpness and gradient-based measures correlate highest with the measured OOD performance. 
Consecutively, \citet{Dziugaite2020InSO} support these findings on sharpness-based and PAC-Bayesian measures as the best-correlated in the similar methodology.

An important application of QE techniques lays in neural machine translation, where avoiding \emph{critical errors} in translation remains an open problem. Such errors deviate the meaning of the translation in a way that may carry health, safety, legal or other implications \cite{specia-etal-2021-findings}. \citet{10.1145/3109480} train a token-level estimator of machine translation output quality concurrently with the neural translation model.
\citet{fomicheva-etal-2020-unsupervised} additionally propose to predict output quality from \emph{entropy of attention activations} of transformer model, but they find this approach not more accurate than the one based on simple output entropy \cite{10.1145/3109480}, or than the MC dropout method.

\subsection{Training Robust Models (RQ2)}
\label{sec:ood-learning}

A problem of training a model that performs well on out-of-distribution (OOD) samples can be found in the literature under the terms of \textit{out-of-distribution generalisation} \cite{Yi2021ImprovedOG}, \textit{domain generalisation} \cite{Gong2021ConfidenceCF}, \textit{distributional robustness} \cite{Meinshausen2018CAUSALITYFA}, or simply \textit{generalisation} \cite{Foret2021SharpnessAwareMF}. The variety of terminology points to the fact that the standards in this branch of research are not yet clearly set. 

Despite imperfect correlations of generalisation measures with measured OOD performance, we find these measures already incorporated in novel training objectives reaching attractive enhancements of model robustness; \citet{pmlr-v40-Neyshabur15} investigate the impact of incorporating norm-based measures into the loss, obtaining generalisation guarantees of $\ell_2$-norm. \citet{Foret2021SharpnessAwareMF} enrich the cross-entropy loss with a complementary component reflecting a sharpness of local optimum, based on a difference to local $\epsilon$. \citet{Keskar2017OnLT} also demonstrate that the sharpness of the objective's optima corresponds to the model's robustness, and flatter optima can also be reached by noising the update steps by smaller training batch size. 

Objective adjustments creatively utilising PAC\discretionary{-}{-}{-}Bayesian measures also confirm reported correspondence of these measures to generalisation. \citet{Hinton2002TrainingPO} proposes a \emph{Product of Experts} (PoE) framework where an ensemble of identical shallow estimators eliminate model-specific biases in a dot product of ensembled outputs, resulting in superior OOD performance. \citet{Sanh2021LearningFO} show an application of PoE eliminating the systematic biases on adversarial NLI data sets. \citet{Dagaev2021ATP} adopt similar approach in debiasing image classification from \emph{heuristical shortcuts}. 
\citet{Utama2020TowardsDN} eliminate model reliance on domain-specific attributes in a two-step process: by \textit{identifying} the biased samples by model over-confidence, and their subsequent \emph{down-weighting}.

Rather than encouraging specific model features, others have investigated the impact of specific \emph{training strategies}, which becomes particularly relevant in multi-step training strategies of LLMs.
\citet{wang-sennrich-2020-exposure} enhance robustness of the translation by fine-tuning for sentence-level Minimum Risk Training objective instead of the common token-level cross-entropy.
\citet{multitask_eliminates_biases} show on adversarial data sets that: a)~longer fine-tuning eliminates model fragility on under-represented samples, and b)~\emph{multitask learning} has a positive impact on transformer generalisation to adversarial data sets. Compliant results are reported by \citet{xie2019unsupervised} with multitask learning for both classification and output consistency to augmented samples, or by \citet{t5} on generative language multitask learning, or in cross-lingual settings by \citet{clark-etal-2019-bert, xlm, lewis-etal-2020-bart}.

Similar results are reported in work addressing dataset biases. \citet{Utama2020TowardsDN, Nie2019AnalyzingCO, Teney2020OnTV} report that addressing only one bias in domain adaptation hurts the model generalisation on other domains. On the other hand, \citet{Wu2020ImprovingQG} find that addressing multiple biases at once can enhance OOD generalisation, although they draw this conclusion from a single domain.

A different branch of work attempts to enhance the robustness by training strategies that work with knowledge of \emph{domain distinction}.
\citet{Gong2021ConfidenceCF} propose to approximately cover the class of \emph{all} possible \emph{target} domains $D_t$ by \emph{source} domains $D_s$ and to learn the calibration of output probabilities from $D_s$ that will allow to \emph{associate} samples of a new target domain $D_t$ to some known $D_s$.
\citet{Yi2021ImprovedOG} propose to use the adversarial framework, learning \textit{indistinguishable} final-layer representation for different domains.

\section{Research Proposal}
\label{ch:objectives}

Following the referenced studies on evaluation and enhancement of the generalisation of neural models, this section outlines directions in measuring and improving robustness of LLMs, respectively.

\subsection{Estimating Model Robustness (RQ1)}
\label{ref:evaluative_measures_objective}

Recently, the measures of generalisation of neural networks struck increasing attention \cite{Jiang2020FantasticGM, Dziugaite2020InSO}. However, none of the referenced studies evaluates the measures on the case of LLMs. Especially within a standard \emph{pre-training + fine-tuning} framework of modern NLP applications, quality of the measures might differ compared to the experiments on relatively small convolutional networks trained for image classification from scratch.

Hence, we first focus on evaluating the established generalisation measures, such as the ones based on spectral complexity, variance of gradients or sharpness in the case of pre-trained LLMs. A~major challenge is to scale such experiments to a representative evaluation framework covering a~broad set of tasks, domains, and model types. For instance, other training parameters will likely impact the metrics' quality; such covariates will have to be identified and controlled. However, even extensive evaluation will likely fail to identify some of such covariates; Due to this reason, we will delimit the scope of our results to the estimation and enhancement of robustness \textit{with respect to} the enumerated covariates, even though it contrasts with the methodology of previous work.

We will give preference to the generalisation measures that correspond to linguistic and semantic language properties, as the practical deployment of such measures in evaluation also addresses a desire for enhancing \emph{interpretability} of the LLMs' behaviour. Instances of linguistically-motivated measures can be a \textit{largest common ancestor} between the parse trees of reference and hypothesis of generative model, or a coherence of output of discriminative model when a negation is introduced in the input.

In the evaluation of robustness of generative LLMs, we will prioritise \emph{token-level} measures over conventional segment-level ones such as BLEU, as incorporating accurate token-level measures in training objectives could complement the classic token-level cross-entropy loss in sequence-to-sequence objective with its specific flaws, such as \textit{exposure bias} \cite{wang-sennrich-2020-exposure}.

The evaluation methodology will closely follow the one of \citet{Dziugaite2020InSO}, which reflects on a correlation of the measure with the measured OOD performance. If these measures reach high correlations, they might be applied directly in training regularisation or model selection. Even in cases of measures not reaching a high correlation, these can still bear the potential to improve model robustness \cite{Foret2021SharpnessAwareMF}.

\subsection{Training Robust Models (RQ2)}
\label{sec:objectives_objective}

Following the referenced examples adjusting training objectives with accurate generalisation measures (§\ref{sec:ood-learning}), e.g. norm-based measures \cite{pmlr-v40-Neyshabur15}, PAC-Bayesian measures \cite{Sanh2021LearningFO,Dagaev2021ATP,Utama2020TowardsDN}, or sharpness measure \cite{Foret2021SharpnessAwareMF}, we will use the accurate generalisation measures of LLMs (§\ref{ref:evaluative_measures_objective}) as \textit{regularizers} and \textit{complementary objectives} of the training.

\citet{Locatello2019ChallengingCA} theoretically prove that full distributional robustness is not possible without an \textit{explicit} exposition of both the data and the model biases. Recently, \citet{Bengio2020A} theoretically and empirically demonstrated that the model could \textit{utilise} data biases to expose the underlying causal structure of the data in an experiment where such a structure is preliminarily known.

We will introduce training objectives that expose domain-specific data biases to the model in more explicit ways. 
The most direct approach is to complement the task-specific objective with another objective of distinguishing the domain(s) of origin. 
The domain-distinctive objective can shape a form of a binary classifier or a similarity loss of selected model representations (e.g. KL-divergence \cite{Kullback1951OnIA}).

We will investigate the impact of the \emph{pre-training}, and \emph{fine-tuning} objectives on the model's eventual robustness over multiple application tasks, domains and architectures, in a methodology similar to the \emph{generalisation measures} evaluation of \cite{Dziugaite2020InSO}.

Additionally, we will \emph{replace} or \emph{complement} the objectives of generative LLMs with token-level measures well-correlated with the OOD performance and compare the resulting models with computationally-expensive sentence-level objectives optimising the measures such as BLEU as their objectives. 

In the case of discriminative models, we will evaluate robustness to surface-level heuristics using adversarial datasets like HANS \cite{McCoy2019RightFT}, or PAWS \cite{Zhang2019PAWSPA} designed to expose the commonly-learnt biases of LLMs.
For generative LLMs, we will evaluate a performance of the model on domain(s) \emph{different} from the training domain; for instance, we will train a translation model on \textit{subtitles} parallel corpus and evaluate on a domain of \textit{news articles}. We will also evaluate the trained model(s) for its inclination to \emph{critical errors} as a probability of generating a translation containing a severe error \cite{specia-etal-2021-findings} in \emph{enforced} generation.

\section{Conclusion}

Our work outlines potential directions in enhancing distributional robustness of LLMs to mitigate a~performance drop under distribution shift.
We survey and identify accurate generalisation measures (§\ref{sec:evaluation}) and find multiple studies demonstrating that utilisation of these measures in the training objectives positively impacts model robustness (§\ref{sec:ood-learning}). 

Following this observation, we propose to identify generalisation measures best-suitable for LLMs (§\ref{ref:evaluative_measures_objective}) and outline ways how to utilise these measures in the training process. Additionally, we identify a set of other methods reported to enhance OOD performance of LLMs that we propose to compare to in the outlined methodology for evaluating generalisation measures. 

Similarly, we propose methodologies for robustness estimation of both generative and discriminative LLMs (§\ref{sec:objectives_objective}); These methodologies are based on the model's quality assessment on the domains covered by the explicitly enclosed set of perturbations or adversarial biases.

\bibliography{my_personal}
\bibliographystyle{acl_natbib}

\end{document}